\documentclass[1p]{elsarticle}

\usepackage{lineno}
\modulolinenumbers[5]

\journal{}

\biboptions{authoryear}

\makeatletter
\def\ps@pprintTitle{%
  \let\@oddhead\@empty
  \let\@evenhead\@empty
  \def\@oddfoot{\reset@font\hfil\thepage\hfil}
  \let\@evenfoot\@oddfoot
}
\makeatother

\usepackage{amssymb}
\usepackage{amsmath}
\usepackage{multirow}

\usepackage{subfigure}

\usepackage{url}

\usepackage{array,tabularx}

\usepackage{algorithm}
\usepackage[noend]{algpseudocode}

\usepackage{adjustbox}
\usepackage{makecell}

\begin{document}

\begin{frontmatter}

\title{Supervised learning and tree search for real-time storage allocation in Robotic Mobile Fulfillment Systems}

\author[label1,label2]{Adrien Rim\'el\'e\corref{cor1}}
\cortext[cor1]{Corresponding author}
\ead{adrien.rimele@polymtl.ca}

\author[label5]{Philippe Grangier}
\author[label1,label3]{Michel Gamache}
\author[label1,label2]{Michel Gendreau}
\author[label1,label2]{Louis-Martin Rousseau}

\address[label1]{Department of Mathematics and Industrial Engineering, Polytechnique Montreal}
\address[label2]{CIRRELT, Research Centre on Enterprise Networks, Logistics and Transportation}
\address[label3]{GERAD, Group for Research in Decision Analysis}
\address[label5]{IVADO Labs}

\begin{abstract}

A Robotic Mobile Fulfillment System is a robotised parts-to-picker system that is particularly well-suited for e-commerce warehousing. One distinguishing feature of this type of warehouse is its high storage modularity. Numerous robots are moving shelves simultaneously, and the shelves can be returned to any open location after the picking operation is completed. This work focuses on the real-time storage allocation problem to minimise the travel time of the robots. An efficient -- but computationally costly -- Monte Carlo Tree Search method is used offline to generate high-quality experience. This experience can be learned by a neural network with a proper coordinates-based features representation. The obtained neural network is used as an action predictor in several new storage policies, either as-is or in rollout and supervised tree search strategies. Resulting performance levels depend on the computing time available at a decision step and are consistently better compared to real-time decision rules from the literature.

\end{abstract}


\end{frontmatter}


\section{Introduction}
\label{sec:introduction}

E-commerce has not only revolutionised access to goods, it has fundamentally changed the entire Business-to-Consumer market segment. Its growth in the last decade has been steady, with an annual increase of approximately 15\% in the US market. The year 2020, exceptional in all regards, reached new summits with a stunning 44\% increase in US e-commerce sales \citep{Ali2021US360}. E-commerce market share now represents 21.3\% of all retail shares, totalling \$861B in the US alone. Specific features of online sales, as well as fierce competition between retailers, put intense pressure on the supply chain, including the warehousing systems.

In response to increasing demand, new warehousing systems have appeared in recent years. One of them is the Robotic Mobile Fulfillment System (RMFS), also referred to as a rack-moving mobile robot based warehouse, semi-automated storage system, or Kiva warehouse \citep{Wurman2008CoordinatingWarehouses, Boysen2019WarehousingSurvey}. In this parts-to-picker warehouse, a fleet of small robots is tasked with retrieving storage shelves and bringing them to picking stations where human operators pick items to fulfil orders. Kiva Systems first introduced the RMFS in 2006 before being bought by Amazon in 2012 and re-branded as Amazon Robotics. Similar systems have been introduced by other companies, including the Zhu Que robots (Alibaba), the Quicktron robots (Huawei), the CarryPick system (Swisslog), and others \citep{Banker2018NewAutomation, Boysen2019WarehousingSurvey}. Sales of warehouse robots are expected to reach \$30.8B by 2022 for around 900,000 robot units sold per year \citep{Valle2021OrderEnvironment}. In an RMFS, the items inventory is stored in mobile shelves that robots can lift and bring to a picking station. The robots typically perform \emph{dual command cycles}, which consist of immediate successions of storage and retrieval tasks. Interestingly, the robots have to move along the aisles when loaded but can pass under the stored shelves when unloaded to avoid traffic. For more technical information about RMFSs, the interested reader is referred to \cite{Wurman2008CoordinatingWarehouses, Azadeh2017RobotizedOpportunities} and \cite{Boysen2019WarehousingSurvey}. A distinguishing feature of an RMFS is its high storage modularity, since the robots operate across the entire storage area simultaneously and can select any free storage location when returning a shelf. This presents great opportunities to continuously adapt the layout of the warehouse in order to accommodate new incoming orders.

This work focuses on the real-time storage allocation problem to minimise the average travel time of the robots, which is inversely proportional to the maximum order throughput value. In this problem, stochastic incoming orders are continuously revealed in the list of orders to fulfil, and a storage location must be chosen as soon as a robot is released by an operator at a picking station. The contribution of this work is to propose a new approach based on supervised learning to this real-time problem. First, an efficient -- yet computationally expensive -- Monte Carlo Tree Search (MCTS) method is proposed to generate high-quality experience offline. This experience is later used to learn a policy that performs better than baselines from the literature while being just as fast to deploy. The features representation, using the shelves' spatial coordinates in the storage area, is crucial to successful supervised learning by a neural network. With additional computing time at the decision step, the learned policy is further improved by a rollout strategy and a newly introduced supervised tree search method, which provide significant performance gains. 

The remainder of this paper is organised as follows. Section~\ref{sec:literature_review} presents a literature review on RMFSs. Section~\ref{sec:problem definition} describes the problem and the different assumptions that are made. Section~\ref{sec:methodology} explains the methodology, starting from the offline MCTS method, followed by the supervised learning of a storage policy, and enhancements with rollout strategy and a new supervised tree search method. Section~\ref{sec:simulation_study} presents a simulation study of the proposed methods compared to baselines from the literature. Finally, Section~\ref{sec:conclusion} draws conclusions and discusses future research perspectives.

\section{Literature review}
\label{sec:literature_review}

While the deployment of RMFSs is relatively recent, research on the topic has its origins in the field of Automated Storage and Retrieval Systems (ASRSs) starting in the 1970s \citep{Roodbergen2009, Gagliardi2012ModelsReview}. In its most typical implementation, an ASRS corresponds to a juxtaposition of static racks, each with a single, aisle-captive crane retrieving bins. Just as in RMFSs, an ASRS's crane performs dual command cycles. The fact that RMFSs use a fleet of robots that can simultaneously access the entire storage area significantly changes the operational decision-making problems. Due to the unique challenges of RMFSs, specific literature has developed on the topic in the last few years.

\subsection{Analytical models}
Several papers develop analytical models to quickly estimate key performance indicators. 
\cite{Lamballais2017EstimatingSystem} propose semi-open queueing network models to estimate the maximum order throughput, average order cycle time, and robot utilisation, depending on layout decisions and robot zoning strategies. They demonstrate the accuracy of their model compared to results from simulations, and note that the throughput is more strongly influenced by the location of the picking stations than by the dimensions of the storage area. 
\cite{Zou2017AssignmentRetailers} present a semi-open queueing network to evaluate decision rules to assign robots to picking stations, and show that their rule performs better than a random assignment. 
In \cite{Yuan2019}, the authors propose a fluid model to assess the performance of velocity-based storage policies in terms of travelling time to store and retrieve shelves. They find that when the items are stowed randomly within a shelf, a 2-class or 3-class storage policy reduces the travelling time anywhere from 6 to 12\% (a class regroups containers by demand rate, and classes of higher demand are typically localised closer to the picking stations). Additionally, they find that if items are stowed together based on their demand velocity, the loaded travelling times can be reduced by up to 40\%. 
\cite{Roy2019} use a closed queueing network model to evaluate the impact of allocating robots exclusively to either retrieval or replenishment tasks, or a combination of both. They conclude that the third option reduces picking time by 30\% but increases replenishment time by a factor of three. They also study storage allocation to multiple classes and find that sending robots to the least-congested class gives similar performance compared to a dedicated class policy. 
\cite{LamballaisTessensohn2020InventorySystems} propose a semi-open queueing model to optimise the number of shelves per item type, the ratio of the number of picking stations to replenishment stations, and the replenishment level per shelf. They conclude that there is a significant throughput gain when: the same item is stored in multiple shelves, the ratio of picking stations to replenishment stations is optimised, and a shelf is replenished before it is empty.

\subsection{Scheduling and task allocation}
Other papers focus on scheduling the processing of orders at the picking stations. \cite{Boysen2017Parts-to-pickerEnvironment} present a Mixed Integer Program (MIP) to minimise the number of shelf visits by synchronising the orders processing and shelves' arrival at a picking station. They propose several heuristic methods based on dynamic programming and simulated annealing to solve the model. Through simulations, they show that their approach can halve the robot fleet size in the best case. 
\cite{Gharehgozli2020RobotSystem} address the problem of scheduling a batch of orders for a single robot to minimise the travelling time. They consider two cases where a shelf either must be stored back at its exact same storage location or can be returned to a set of locations. They demonstrate the efficiency of their adaptive large neighbourhood search method and show that the scheduling improves travelling time by 24\% compared to a random order.
\cite{Valle2021OrderEnvironment} present a mathematical model and matheuristic solving methods to simultaneously allocate a batch of orders and shelves to multiple pickers. They also propose a model to sequence the shelves' visits to the picking stations.
\cite{Bolu2021AdaptiveWarehouse} propose a task planning approach that aims at minimising the number of shelves needed to fulfil a batch of orders. Then, they dynamically adapt the system by selecting tasks for the picking stations and the robots. Through an extensive simulation study, they demonstrate their method's capacity to significantly reduce the completion time and maintain a high pile-on (i.e., the number of items retrieved from a single shelf visit).

From a different perspective, \cite{Zhang2019ASystem} define the problem of allocating robots to picking tasks as scheduled at the picking stations. They formulate a resource-constrained scheduling problem and propose a solving approach using a serial schedule generator and a genetic algorithm. They show that their method obtains better results than commonly used rule-based scheduling methods. 
\cite{Xie2021IntroducingSystems} propose an MIP to assign shelves to stations and orders to stations simultaneously. They also consider splitting multi-line orders for picking at different stations. They use the discrete event simulator RAWSim-O from \cite{Merschformann2018RAWSim-O:Systems}, which realistically reproduces operations in an RMFS, to compare their results with a sequential allocation. With split orders, they obtain a 46\% increase in throughput.
Using this simulation framework, \cite{Merschformann2019DecisionSystems} assess combinations of decision rules for operational problems such as picking and replenishment station assignment, pod selection, and storage assignment. In particular, they note the importance of careful selection for each application, the significant influence of the picking station assignment rule, and the high cross-dependencies between some rules.
\cite{Rimele2020} present a stochastic mathematical modelling framework that formalises real-time decision-making. Presented as a stochastic dynamic program, it considers most operational decisions from order sequencing to storage allocation, with stochastic incoming orders. Its main objective is to set a framework to stimulate the development of advanced methods for real-time decisions in RMFSs.

\subsection{Storage allocation}
Regarding storage allocation specifically, \cite{Krenzler2018DeterministicSystems} present a deterministic MIP model aiming to minimise the shelves' storage and retrieval times. They consider that the shelves’ visit times at the picking station are known and assume that the waiting queues are always full, which frees up more storage locations. They present a variety of solving methods and demonstrate their effectiveness compared to a random storage policy. 
\cite{Weidinger2018StorageWarehouses} present an MIP model in the form of an interval scheduling problem to minimise the loaded travel time. They also consider that the visit time of each shelf at the picking station is known. With this assumption, the model enforces that two storage allocations of shelves can occupy the same location only if the corresponding storage intervals do not overlap. They present an Adaptive Large Matheuristic Search solving method and obtain conclusive results in terms of loaded and complete travel times, compared to standard decision rules. They note the good performance and practical aspect of the Shortest Leg policy (referred to as Shortest Path), which only increases the travel time by 3.49\% and does not require any batching of orders.
\cite{Mirzaei2021ThePerformance} use product affinity (i.e., the probability of products being ordered together) and turnover rate to jointly assign products to shelves, and shelves to locations, to minimise the retrieval time with an integer model. They show that their model reduces retrieval time by up to 40\% compared to traditional full-turnover and class-based storage policies.
\cite{Rimele2021E-commercePolicy} use Reinforcement Learning techniques to learn a zone-based storage policy that minimises the travel time of robots with one picking station. Depending on the distribution of orders, they improve the Shortest Leg (SL) policy's performance by 3.5 to 5\%. They also propose a lookahead rollout strategy that uses already-revealed orders to simulate the near horizon. Compared to SL, this approach generates results that are 7.5\% better on average.

This paper is presented as a continuation of \cite{Rimele2021E-commercePolicy}, as its objective is to learn an efficient real-time storage policy. However, the methods proposed here select exact storage locations instead of zones and rely on a significantly different learning approach. The following section defines the problem we consider and the underlying assumptions.

\section{Problem definition}
\label{sec:problem definition}

This work focuses specifically on storage allocation when a robot has to return a shelf to the storage area, with the objective of minimising the average travel time. Travel time corresponds to the complete cycle of a robot within the storage area; it includes the storage time from the picking station to the storage location, the interleaving time from the storage location to the next retrieval location, and the retrieval time from the retrieval location to the picking station. \cite{Rimele2020} present a model as a Markov Decision Process where a step corresponds to a state of the system when a robot is available at a picking station, and a decision has to be made between two types of tasks: a regular storage task or an opportunistic task. The latter corresponds to a situation where the shelf that is currently being carried by the robot can be used to fulfil another order instead of returning to the storage area. A regular storage task consists of selecting a storage location that is currently available. When a robot has stored a shelf, it retrieves the next assigned shelf and brings it to a picking station's waiting queue. Orders are stochastic and can enter the list of orders to fulfil at any moment, with any priority level (an Amazon \emph{Prime} order would, for instance, be pushed to the top of the list). As such, the system needs to adapt to new orders continuously (no batching). While the future list of orders is uncertain at the decision step, the list of already-revealed orders is available. Several assumptions are made and listed below.

- A1. Only one picking station is considered.

- A2. The arrival of new orders is stochastic, but travelling and processing times are deterministic.

- A3. Opportunistic tasks, as described above, are automatically enforced when possible. 

- A4. Since the focus is on the storage allocation problem, incoming orders are directly associated with a shelf. The sequencing is updated according to the order's deadline, with the condition that the corresponding shelf is not currently being carried by another robot.

- A5. Congestion and path optimisation for the robots are ignored.

- A6. Notions of stock levels and replenishment tasks are not considered.

- A7. Decisions have to be made swiftly, but we have access to a simulator that can be used offline to accumulate valuable experience. The simulator has the ability to copy its current state to run alternative future outcomes (with the essential condition of not revealing unknown information).

Following these assumptions about the real-time storage allocation problem, Section~\ref{sec:methodology} presents the different methods developed in this work to tackle the problem.

\section{Methodology}
\label{sec:methodology}

This section presents different methods, each with its advantages and drawbacks in terms of performance and computing time. All of these methods use the same base policy, which is learned offline from an expert system, namely an MCTS algorithm. Section~\ref{sec:MCTS} describes how we adapted an MCTS algorithm to the storage allocation problem. Using the experience accumulated from this MCTS algorithm, Section~\ref{sec:learning_policy} explains how to learn a fast and efficient policy. Section~\ref{sec:enhancement_learnt_policy} improves the learned policy's performance by using rollout strategies and a supervised MCTS. Finally, Section \ref{sec:deep_mcts} presents a new algorithm that we call Supervised Tree Search (STS), which leverages the learned policy by exploring the search tree differently.

\subsection{Monte Carlo Tree Search algorithm}
\label{sec:MCTS}

\begin{figure}
\centering
\includegraphics[width=120mm]{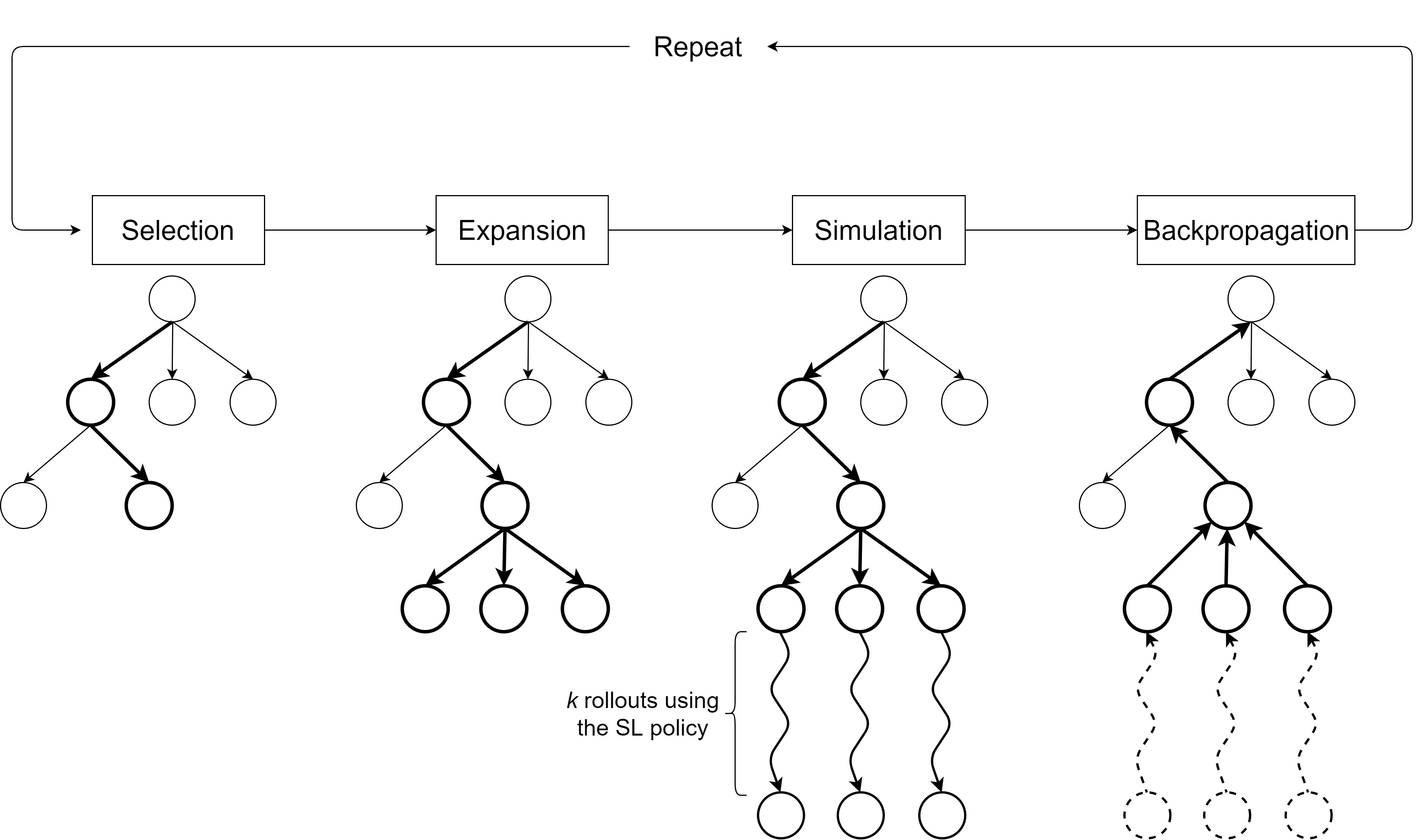}
\caption[MCTS]{MCTS - The Selection step uses a selection rule (e.g., UCT) to select nodes (actions) within the tree until a leaf node is reached. The Expansion step creates one child or several children nodes from the current leaf node. The Simulation step runs simulations (trajectories) from a newly-created node using a fast base policy, either until the end of the game or for a given number of steps. Finally, the Backpropagation step updates the return value of the visited nodes based on the rewards revealed along the trajectories.}
\label{fig:mcts}

\end{figure}

Monte Carlo Tree Search is a method to optimise sequential decisions, relying on sampling to build an exploration tree, in which a node represents an action. It was initially designed for finite and binary decision games, but it is applicable -- with adjustments -- to any sequential decision problem \citep{Browne2012AMethods}. Using random sampling to approximate decisions' values from Monte Carlo methods, \cite{coulom2007} first proposed a Monte Carlo Tree Search algorithm. The idea is to run simulations until the end of the game and to use experience gained to incrementally build a tree, one node at a time. Within the tree, the nodes, or decisions, are selected based on the probability that they will perform best. The rest of each simulation makes decisions randomly, until a terminal state is reached. As depicted in Figure~\ref{fig:mcts}, a general MCTS iteration contains the following four steps: \emph{Selection}, \emph{Expansion}, \emph{Simulation} and \emph{Backpropagation}. \cite{Kocsis2006BanditPlanning} adapt a selection rule from the multi-armed bandit domain, namely the \emph{Upper Confidence Bound (UCB1)} of \cite{Auer2002Finite-timeProblem}, to MCTS and call it \emph{Upper Confidence Bounds for trees (UCT)}. Inside the tree, UCT selects the child node $i$ that maximizes $\bar{V_i} + 2 c \sqrt{\frac{ln\left ( N \right )}{n_i}}$, where $\bar{V_i}$ is the running average return value of node $i$, $N$ is the number of visits of the parent node, $n_i$ is the number of visits of node $i$, and $c$ is the exploration parameter (the higher the value of $c$, the greater the likelihood of selecting a less-visited node). 

Figure~\ref{fig:mcts} presents an iteration of the MCTS algorithm as used in this work. Unlike win-or-lose games, our system evolves in an infinite horizon and receives continuous rewards (the travel time of the robot) at every step. To adapt MCTS in our case, we first truncate the horizon after $h$ steps. Second, to maintain the current average return value between 0 and 1, and to keep the two terms within the same order of magnitude, we propose to adapt the UCT equation with Eq.~\ref{eq:UCT}, where $\bar{V}$ represents the running average return value of the parent node. 

\begin{equation}
    \label{eq:UCT}
    \textup{UCT} = \underset{i}{\textup{argmax}} \; \frac{1}{\bar{V_i}} + \frac{c}{\bar{V}} \sqrt{\frac{ln\left ( N \right )}{n_i}}
\end{equation}

Note that $\bar{V_i}$ would only be equal to 0 if the decision of node $i$ corresponds to an opportunistic task. Because we systematically enforce such tasks, UCT does not apply. For all other tasks, a cycle takes at least 1 second, so that $\frac{1}{\bar{V_i}}$ is always between 0 and 1 in Eq.~\ref{eq:UCT}.

In our system, when MCTS is executed, a sequence of decisions results in a deterministic output. This is explained by the fact that only revealed orders are considered in the trajectories, and travelling and processing times are assumed to be deterministic. This observation justifies two notable changes from a common MCTS. First, as depicted at the \emph{Expansion} step in Figure~\ref{fig:mcts}, and as described in \cite{Browne2012AMethods}, a full node set expansion is performed. If $k$ sibling nodes would normally be visited, this saves simulation time corresponding to the number of steps inside the tree. Indeed, this sub-trajectory would result in the same, redundant outcome before each child's first visit. Second, like in \cite{Bjornsson2009CadiaPlayer:Player}, at each node, we track not only the current average return value but also the best encountered simulation return value. The average value is still used for the selection step to explore more promising areas of the tree, but the actual final choice of action will select the node with the best encountered value. Empirically, these design choices gave the best results in our simulation study.

Finally, we use the Shortest Leg (SL) policy as the base policy at the \emph{Simulation} step, instead of a Random policy. Indeed, as in \cite{Rimele2021E-commercePolicy}, the SL storage policy is a relatively efficient and quickly deployable policy.

\subsection{Learning a policy}
\label{sec:learning_policy}

While the MCTS algorithm presented above is capable of excellent performance to minimise the average cycle time, its performance is strongly dependent on the number of trajectories it runs at every decision step. The number of trajectories is directly correlated to the running time, limiting the opportunity for a satisfactory real-time deployment. Instead, we propose to use MCTS offline, when computing time is not an issue, to accumulate high-quality experience. Different versions of MCTS, corresponding to different truncated horizons $h$, are used to generate datasets of experiences that we refer to as \emph{MCTS(h)}. The content of these datasets is described below.

\subsubsection*{Features representation}

The complete state of the system at a decision point contains a tremendous number of variables \citep{Rimele2020}, which would make any type of learning particularly difficult. Instead, we propose to define a Partially Observable Markov Decision Process (POMDP), where the observable state of the system corresponds to a vector of a limited number of key features. These features are chosen based on the expectation that they would provide sufficient information to make good decisions.

\begin{figure}
\centering
\includegraphics[width=70mm]{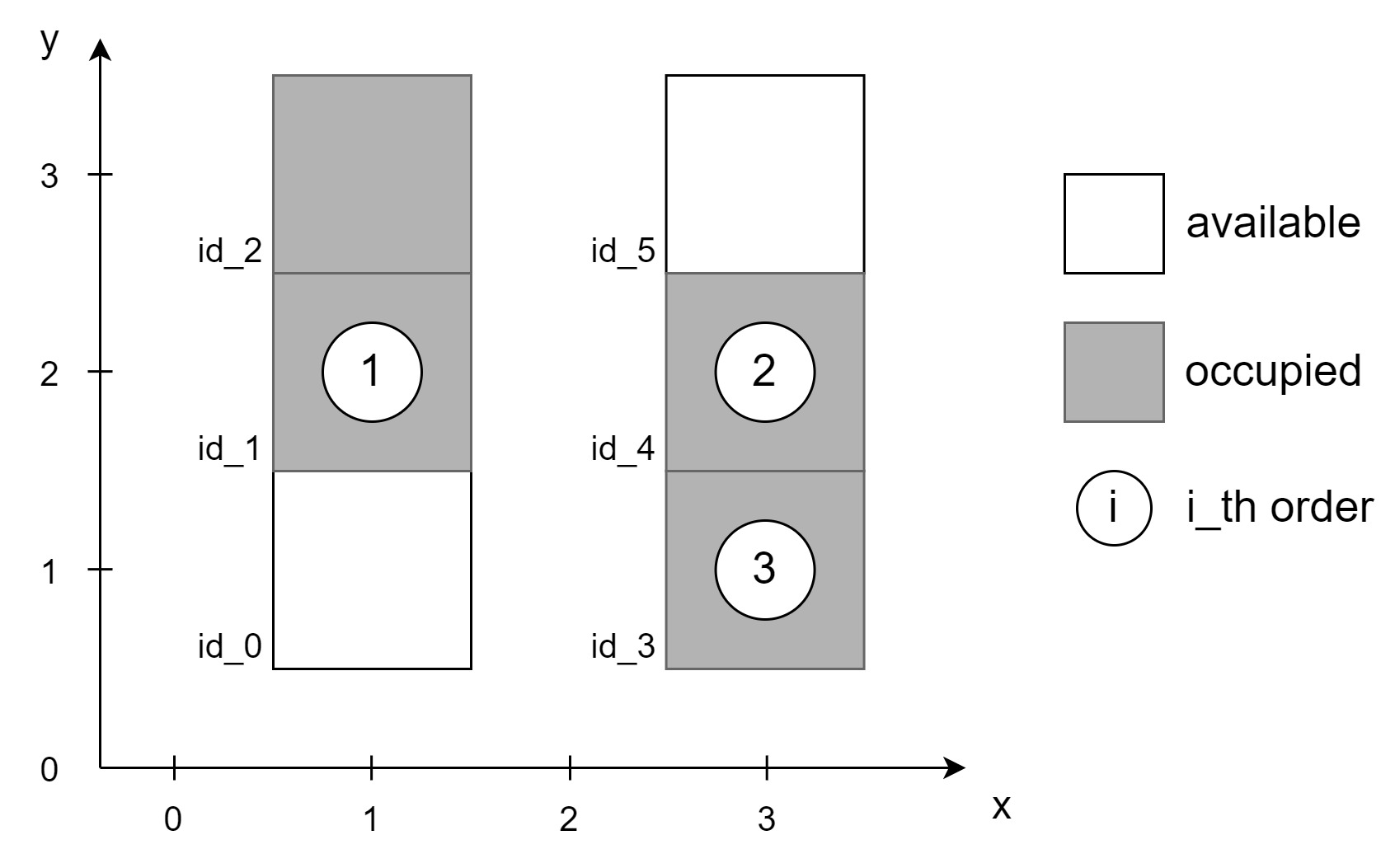}
\caption{Simple features representation illustration}
\label{fig:features_representation}
\end{figure}

Figure~\ref{fig:features_representation} presents a simplified example of a warehouse containing six storage locations, four stored shelves and three waiting orders. Two locations are currently available, and the sequence in which the revealed orders must be retrieved is known and written on the corresponding shelves.

The extracted features are two-fold. First, we represent which locations are available. If the warehouse is full (i.e., the same number of shelves and locations) and uses $r$ robots (four in the example), the maximum number of available locations at any time step is $r$. These locations are ordered by increasing id and represented in the features vector by their $(x,y)$ coordinates in the warehouse. If less than $r$ locations are available, a mock location of coordinates $(-1,-1)$ is used to fill the remaining reserved features. Similarly, the next $h$ orders are each represented by the corresponding shelves' coordinates. In the example, the resulting features vector would be: 
$$[ \: \underset{\text{available locations}}{\underbrace{(1,1)\;  ,\; (3,3)\;  ,\; (-1,-1)\;  ,\; (-1,-1)\;  }},\underset{\text{next }h \text{ orders' locations}}{\underbrace{\; (1,2)\;  ,\; (3,2)\; , \; (3,1)}}]$$
Using coordinates instead of location ids is preferable for generalisation during the learning. In general, if $r$ denotes the maximum number of available locations (here equivalent to the number of robots) and $h$ the number of revealed orders considered, the features vector contains $2(r+h)$ features. Finally, a features vector is saved in the dataset at every decision step, along with the id of the storage location decided by the MCTS method. Note that other features representations are worth considering, but this one gave the best results in terms of classification accuracy and computing time (see next paragraph and Section~\ref{sec:simulation_study}).

\subsubsection*{Learned Policy}

A dataset MCTS(h) can now be used to learn a policy. This learning consists in having a classifier that takes as input a features vector, as previously described, and predicts the probabilities of each storage location to be selected. This classifier is denoted as $f: \mathbb{R}^{n} \mapsto \left [ 0, 1 \right ]^{\left |\mathcal{L}  \right |}$, where $n$ is the number of features and $\mathcal{L}$ the set of storage locations. From this classifier, a \emph{Learned Policy (LP)} can be defined as: $LP(X) = \underset{l}{\textup{argmax}}\left ( f(X)_l \mid l \in \mathcal{L}^{\textup{avail}} \right )$, where $X \in \mathbb{R}^{n}$ is a features vector, $f(X)_l$ refers to the $l^{\textup{th}}$ element (location) of $f(X)$, and $\mathcal{L}^{\textup{avail}}$ is the subset of the locations that are currently available. In other words, at every decision step, LP extracts a features vector from the current state of the system, and selects the available storage location with the highest predicted probability according to the classifier.

\subsection{Enhancement of the Learned Policy}
\label{sec:enhancement_learnt_policy}

One advantage of a policy like LP is that it is extremely fast to deploy since it immediately selects an action based on the current state representation. However, by construction (supervised learning on a heuristic output) and due to the nature of the system (stochastic and partially observable), such a policy is necessarily sub-optimal. As described in \cite{Bertsekas2019ReinforcementControl}, a lookahead strategy can be employed to take better advantage of a base policy. As used by \cite{Rimele2021E-commercePolicy} in a similar setting, a simple yet efficient approach is to perform a one-step lookahead with truncated rollouts. This approach runs simulations using already-revealed orders on every feasible action from the current state and deploys the base policy over $h-1$ more steps. When the simulation is completed, the initial action that resulted in the best outcome is selected. We refer to the resulting policy as \emph{Base Policy+rollouts}; for instance, LP+rollouts.

A natural extension of this approach is to consider multi-step lookaheads, for which MCTS can be seen as a special construction method. The Learned Policy can be used to guide the Selection step in MCTS. \cite{Silver2016MasteringSearch.} in their famous AlphaGo program use a selection rule that combines a state value estimate and a policy network prediction. Here, we propose to only use the policy network prediction, but to sample the action based on this value instead of taking the argmax. As such, if the parent node presents a features vector $X$, the selection rule is:

$\textup{Supervised Selection}(X) = \textup{Sample}\left ( \left \{ a \in A(X)\right \}, Prob(a)=\frac{pred(a)}{n_a + 1}  \right )$, where $A(X)$ denotes the feasible actions from state $X$, and $pred(a)$ represents the filtered prediction from the Learned Policy: $pred(a)=\frac{f(X)_a}{\sum_{a' \in A(X)} f(X)_{a'} }$.

Although it performs well (see Section~\ref{sec:simulation_study}), the construction of a supervised MCTS may be limited by the lack of computing time for real-time deployment. First and fundamentally, MCTS iteratively builds a search tree running numerous trajectories, which takes time. As a result, it may not branch very deep without a significant number of trajectories. Second, while a regular MCTS uses a fast heuristic rule to select nodes, our supervised MCTS uses a machine learning predictor. Each prediction is costly, mainly because the predictions need to be made sequentially, not in batches. Considering the valuable insights provided by the action classifier, we feel that another approach could be more beneficial. We propose such an approach below.

\subsection{Supervised Tree Search algorithm}
\label{sec:deep_mcts}

This algorithm's objective is to deeply explore the search tree with only a limited number of trajectories. To do so, we consider having an action classifier, here the Learned Policy, in which we can put significant trust.

\begin{figure}
\centering
\includegraphics[width=130mm]{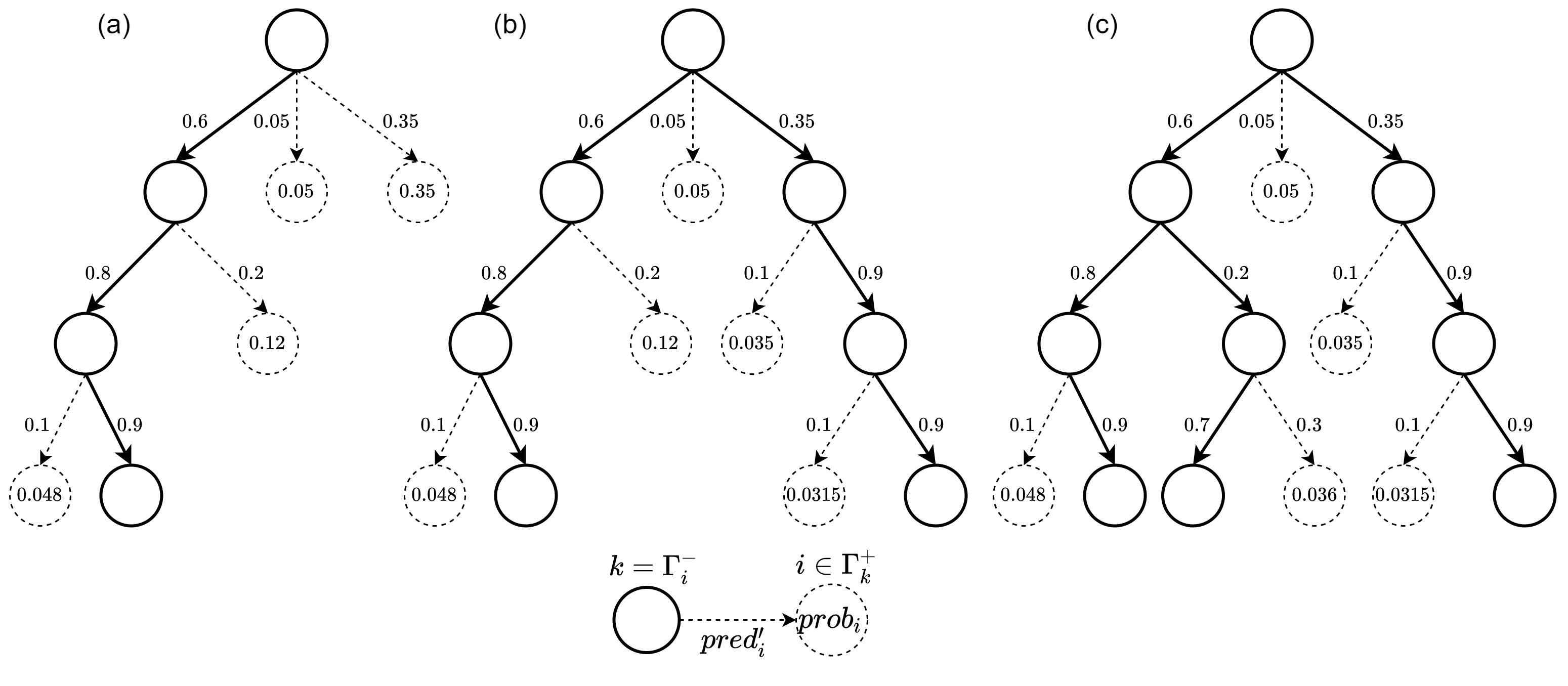}
\caption[Supervised Tree Search]{Supervised Tree Search - Step (a) corresponds to applying the Learned Policy starting from the initial state. Dashed lines represent the four nodes that were not selected along the trajectory. The value at the center of a node represents its visit probability, while the value beside an arrow represents the corresponding action's predicted value. Step (b) identifies the furthest right node as the open node with the highest visit probability. A new trajectory starts from this node and opens new nodes. In step (c), the node with the highest visit probability has a value of 0.12, and the resulting trajectory leaves one node open.}
\label{fig:deep_mcts}
\end{figure}

The intuitive idea is the following. First, the Learned Policy is run until a maximum depth of $h$ is reached. At each decision point, LP selects the feasible action of maximal prediction value. Along this trajectory, children nodes that were not selected are labelled as \emph{open} nodes. For the following iterations, let us suppose that LP draws actions based on their prediction values instead of acting greedily. In that case, every open node has some probability of being reached by LP. This probability can be calculated as the node's prediction value times the probability of the parent node. Every iteration consists of exploring the search tree starting from the open node with the highest probability. From this node, the new trajectory will apply LP and open new nodes. The process is repeated until the maximum number of trajectories is reached. Figure~\ref{fig:deep_mcts} shows a small example of three iterations of the algorithm.

To allow some exploration, and not simply the exploitation of the action classifier, we propose to transform the prediction values by performing the following operation: $pred'_i=\frac{pred_i+\left (0.5-pred_i \right ) \times \eta}{\sum_{j \in \Gamma_k^+}pred'_j}\; , i \in \Gamma_k^+, \eta \in ]0, 1[ $. This transformation tends to push all values toward equal probabilities, and because it is strictly increasing (when $\eta < 1$), the transformation does not change the predicted values' relative order. In this way, all actions keep a certain probability of being visited, even in the eventuality of the classifier predicting a zero probability. Results from the simulation study were rather invariant to small values of $\eta$. A value of $\eta = 0.1$ is used for the results presented below.

Algorithm~\ref{alg:deep_mcts} formalises the procedure. $\mathcal{N}$ designates the set of open nodes, and $N_0$ is the root node corresponding to the current state. $r_N$ represents the immediate reward received when taking the action leading to node $N$, and $V_N$ is the value of the current best trajectory starting at node $N$. $\Gamma^+_N$ denotes the set of children nodes (feasible actions) of node $N$, and the parent of node $N$ is represented by $\Gamma^-_N$. Finally, $\gamma$ denotes the discount factor applied to future return values.

\begin{algorithm}
\caption{Supervised Tree Search (STS)}\label{alg:deep_mcts}
\begin{algorithmic}[1]
    \Function{STS}{$N_0$}
        \State $\mathcal{N} = \left \{ N_0 \right \}$
        \State $\textup{prob}_{N_0} = 1$
        \While{within the computational budget}
            \State $N_s = \underset{N}{\textup{argmax}}\left ( \textup{prob}_N, N \in \mathcal{N} \right )$
            \State $\Call{Learned\_Policy}{N_s}$
            \State $\mathcal{N} = \mathcal{N}\setminus N_s$
        \EndWhile
        \State \Return $\underset{N}{\textup{argmin}}\left ( V_N, N \in \Gamma^+_{N_0} \right )$
    \EndFunction
    
    \Function{Learned\_Policy}{$N_s$}
        \If{$\textup{depth}_{N_s} < h_{max}$}
            \State $N_g = \underset{N}{\textup{argmax}}\left ( \textup{pred'}_N, N \in \Gamma^+_{N_s} \right )$
            \State $r_{N_g} = \textup{step}(N_s \rightarrow N_g)$
            \ForAll{$N \in \Gamma^+_{N_s} $}
                \State $\textup{prob}_N=\textup{pred'}_N \times \textup{prob}_{N_s}$ 
                \State $V_N = +\infty$
            \EndFor
            \State $\Call{Learned\_Policy}{N_g}$
        \Else
            \State $\Call{backprop}{N_s, r_{N_s}}$
        \EndIf
    \EndFunction
    
    \Function{backprop}{$N, R$}
        \If{$V_{\Gamma^-_{N}} > r_{\Gamma^-_{N}} + \gamma \, R$}
            \State $V_{\Gamma^-_{N}} = r_{\Gamma^-_{N}} + \gamma \, R$
            \If{$\Gamma^-_{\Gamma^-_{N}}\neq \varnothing$ }
                \State \Call{backprop}{$\Gamma^-_{N}, V_{\Gamma^-_{N}}$}
            \EndIf
        \EndIf
    \EndFunction
    
\end{algorithmic}
\end{algorithm}

\section{Simulation study}
\label{sec:simulation_study}

This section presents the results of the various methods presented above. The required computing time differs widely from one method to another, and some parameters, such as the number of trajectories or the time horizon, greatly impact the results. This influence is extensively studied. Before presenting the policies' results, Section~\ref{sec:prelim_results} presents preliminary results from the learning stage that are relevant in order to construct the final policies.

\subsection{Preliminary results}
\label{sec:prelim_results}

\begin{figure}
\centering
\includegraphics[width=80mm]{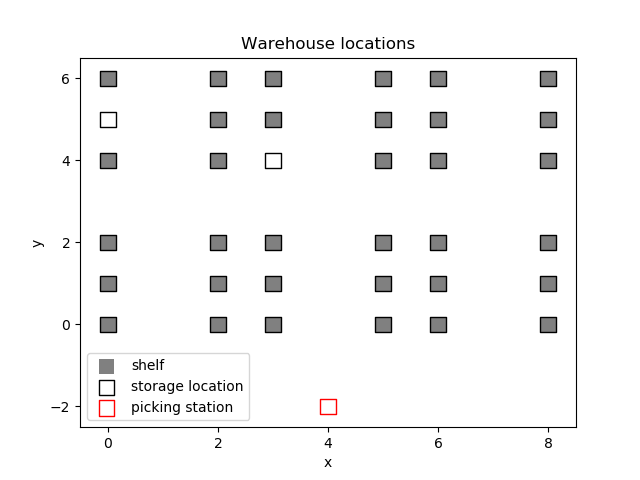}
\caption{Plan view of the storage area}
\label{fig:plan_view}
\end{figure}

Figure~\ref{fig:plan_view} represents the warehouse used in the simulation study. It contains 36 storage locations for as many shelves, and five robots are operating. The other parameters are similar to the ones described in \cite{Rimele2021E-commercePolicy}. A skewed demand distribution is considered, controlled by the cumulative distribution $G(x)=x^s \quad \text{for } 0 < s \leq 1$, where $x$ represents $x\%$ of the shelves and $s$ controls the skewness level. In this study, we set $s=0.7$ for all the tests.

First, four MCTS(h) datasets were generated for h=5, 10, 20, and 30. By trial and error, the exploration parameter $c$ in Section~\ref{sec:MCTS} was set to $1/16$. Each dataset contains 3 million instances, and because there are five robots in the warehouse, the number of features per instance is equal to $2 \times (5 + h)$, so 20, 30, 50 and 70, respectively. Independent from the dataset, the same neural network architecture was used for the supervised learning task. The network is feed-forward and fully connected; it contains five ReLU hidden layers of 100 neurons and a softmax output layer with 36 neurons, one for each storage location. The categorical cross-entropy was used as the loss function, and the Adam optimiser was used. The network was implemented using the Keras library \citep{Chollet2015Keras}. Each training performed 500 episodes with mini-batches of 1024 instances. 

For each dataset, 10\% of the instances were reserved for a test set; the resulting accuracies were 87.16\%, 82.59\%, 77.25\% and 76.46\%. Because of the stochasticity of the transition function and the heuristic nature of the MCTS method, a longer horizon leads to less predictable actions, making the learning task more difficult. However, the longer the horizon, the higher the potential to learn a good-performing policy. Indeed, when testing MCTS with 500 trajectories over different horizons on the test instances in Section~\ref{sec:results}, the corresponding average cycle times are 33.99s, 32.97s, 32.03s and 31.46s. The relationship between learning accuracy and performance creates a trade-off between learning a poor policy better and learning a better policy poorly. The following section will investigate this topic. 

\begin{figure}
\centering
\includegraphics[width=80mm]{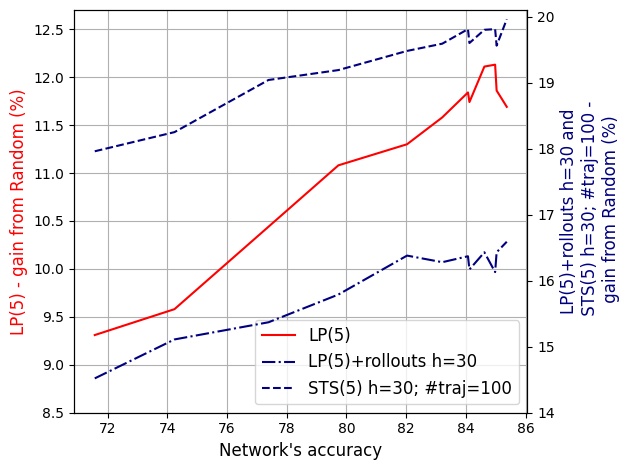}
\caption{Correlation between the network's accuracy and the policies' performance}
\label{fig:correlation}
\end{figure}

While learning from different policies results in different potential performances, one may wonder about the impact of the accuracy when learning from the same policy. Indeed, accuracy is not necessarily proportional to the learned policy's performance, especially if the worst actions are well-learned and avoided. Figure~\ref{fig:correlation} presents a study of the correlation between prediction accuracy and the performance of the learned policies. The y-axis on the left corresponds to the Learned Policy and the y-axis on the right corresponds to LP(5)+rollouts h=30 and STS(5) h=30 \#traj=100. The learning was done on MCTS(5), and the results were generated from 10 instances of 4000 actions each. Smaller neural networks were used to degrade the classification accuracy. The relationship appears to be increasing monotonically, up to until an accuracy of 84\%, which shows that accuracy is a good indicator to guide the network search. Above 84\%, LP's performance seems to be inconsistent, as if a higher accuracy was over-compensated by misclassifications having a stronger impact in the objective function. However, the other learned policies' exploration gives a more consistent performance by reaching some plateau value. This result is of interest in this study when deciding on a network's architecture. Since the network needs to make sequential predictions used in a real-time policy, computing time is critical, and this time is typically dependent on the complexity of its architecture. If two networks present marginal accuracy differences, the simpler, smaller one will be favoured. In particular, we tested a convolutional neural network using a similar input but shaped so that each location corresponds to an element in a matrix. The obtained accuracy was 0.35\% higher than the standard network, but the policies' computing time almost doubled. 

\subsection{Results}
\label{sec:results}

\begin{figure}
\centering
\includegraphics[width=105mm]{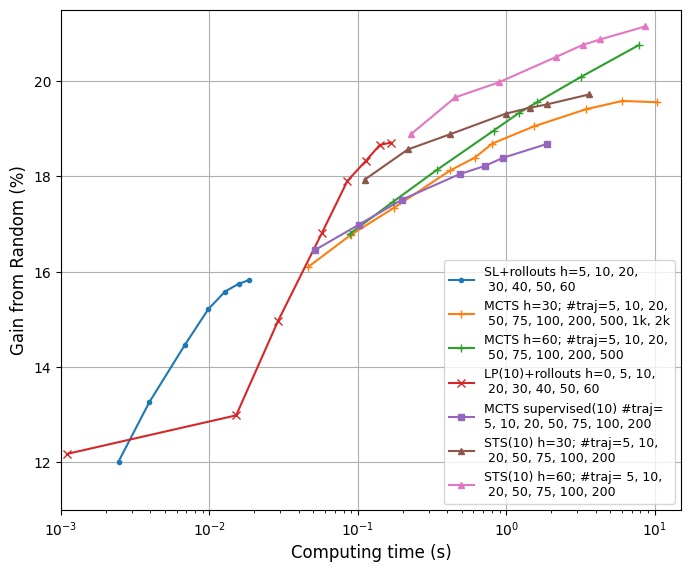}
\caption{Policies' performance}
\label{fig:performance_policies}
\end{figure}

The graph in Figure~\ref{fig:performance_policies} presents the performance of the different storage policies by plotting the performance gain from the Random policy versus the average computing time per decision. All policies were tested over 100 instances of 4000 actions each. The different curves regroup families of policies, which only differ by a parameter value. Each mark along a curve corresponds to a different parameter mentioned in the legend. SL+rollouts and LP+rollouts have their maximum horizon vary up to 60. The curves representing offline MCTS, Supervised MCTS, and STS vary the number of trajectories used in these methods. The horizon of Supervised MCTS is set to 30. Because STS performs better, and for comparison purposes with offline MCTS, STS and offline MCTS have two curves each, corresponding to horizons of 30 and 60.

\begin{figure}
\centering
\includegraphics[width=95mm]{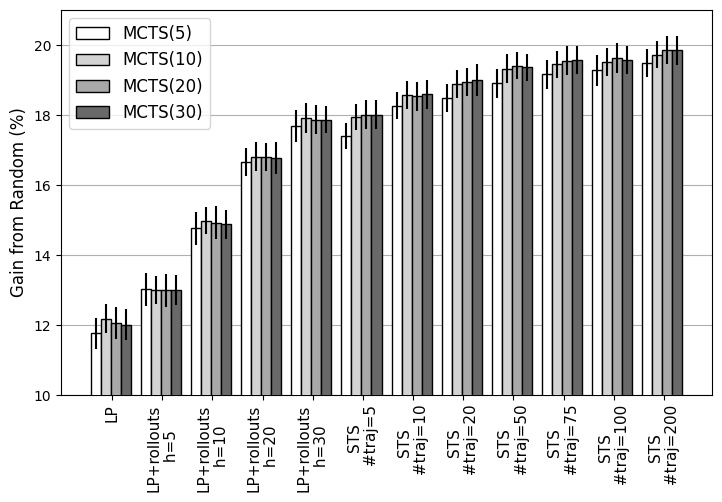}
\caption{Learned policies' performance}
\label{fig:performances_learnt_policies}
\end{figure}

Only the methods learned on MCTS(10) are presented in this graph. This decision is justified by the results presented in Figure~\ref{fig:performances_learnt_policies}, where the policies learned from different datasets MCTS(h) are compared. The general assessment is that the policies improve when learning on longer-horizon datasets (for which the corresponding policies are also better). However, there is a noticeable exception with the Learned Policy that sees its performance decrease after $h=10$. It seems that for $h>10$, the classifiers make more mistakes (as expected with their accuracies), which degrades the policy. However, these classifiers still carry insightful information as they generate better than MCTS(10) results when more exploration is performed, particularly with STS. Overall, because of the good performance and simplicity of MCTS(10), and the marginal gains of MCTS(20) and MCTS(30), we use MCTS(10) for comparisons with the other methods.

For reference, and since it does not appear in Figure~\ref{fig:performance_policies}, SL improves the Random policy by 7.43\% and takes $1.67\mathrm{e}{-4}$ sec. per action. Like SL, which does not perform any exploration, LP performs significantly better with a gain of about 12\% in $1.10\mathrm{e}{-3}$ sec. Using rollouts improves both policies; in particular, rollouts of horizon 30 generate gains of 15.2\% and 17.9\%, and rollouts of horizon 60 generate gains of 15.8\% and 18.8\%. These policies are particularly fast to deploy, with a computing time of less than 0.2 sec./action. 

The family of tree search methods is computationally more expensive, and their computing time is heavily dependent on the number of trajectories run in the exploration. Supervised MCTS appears at first to perform slightly better at similar computing times, and then worse than offline MCTS when more trajectories are run. It seems that the classifier helps find reasonably good trajectories faster, but quickly reaches its limits when more trajectories are allowed. 

STS is the best-performing method overall when more computing time is available. For a horizon of 30, and compared to MCTS with similar computing time, STS runs significantly fewer trajectories (about half) but consistently performs better. With a similar performance, STS runs much fewer trajectories than MCTS. For instance, 10 trajectories with STS results in a similar performance to MCTS with 75 to 100 trajectories, for a quarter of the computing time. The additional time comes from the predictions of the neural network. With only 5 trajectories, STS generates gains of 17.93\% and MCTS 16.10\%. With 200 trajectories, those gains become 19.72\% and 19.05\%, respectively. With a computing time of one second, STS gives gains of about 19.35\% and MCTS 18.80\%.

With a horizon of 60, 5 and 200 trajectories for STS give results of 18.89\% and 21.15\%, while for MCTS the results are 16.79\% and 20.09\%, respectively. With one second of computing time, STS and MCTS generate gains of about 20.10\% and 19.10\%, respectively.

The final choice of policy depends on the available time at the decision-making step. If the time is extremely limited (e.g., 0.001 sec.), LP would be the first choice. With more time (up to 0.1), LP+rollouts appears to be the best choice. If even more time is available, STS gives the best results. Detailed results of all the mentioned policies and their variants are given in \ref{sec:appendixA}.

\section{Conclusion}
\label{sec:conclusion}

This work studied the real-time problem of allocating storage locations in a Robotic Mobile Fulfillment System. The proposed methods rely on learning a policy from the experience accumulated by a Monte Carlo Tree Search algorithm run offline. The neural network used to learn the policy takes as input a features representation of the warehouse's current layout and a list of revealed orders. It predicts the probabilities of allocating a shelf to the storage locations. These probabilities can be used right away to deploy a speedy and well-performing Learned Policy (LP), or used in an exploration strategy. Three such strategies are presented, two of them generating convincing results in terms of performance and computing time. The first one performs lookahead rollouts on each feasible action, using LP as the rollout policy. This approach significantly improves the performance of LP while being very fast to deploy. The second strategy, referred to as Supervised Tree Search, builds a search tree by exploring in priority nodes with the highest probability of being reached by the Learned Policy. This method takes a longer computing time but further improves LP's performance and consistently beats MCTS. The policy choice ultimately depends on the time available for decision-making, and may differ from one warehousing operation to another.

There are several avenues for future research. First, some assumptions made in this work could be relaxed, such as considering several picking stations, notions of robots congestion, and others. A larger warehouse and robot fleet could also be considered to validate the proposed methods' scalability. Another research direction would be to expand on the supervised learning and tree search by integrating it with a Reinforcement Learning algorithm. This could further improve the policy network by leveraging the possibility of learning from higher-quality experiences obtained from the current version of STS.

\section*{Acknowledgments}

We are grateful to the Mitacs Accelerate Program for providing funding for this project. We also wish to gratefully acknowledge the support and valuable insights of our industrial partners, JDA Software and Element AI.

\newpage
\appendix
\section{Detailed results}
\label{sec:appendixA}

\begin{table}[H]
\centering
\label{table:full_results_no_learning}
\caption{Results of the methods without learning}
\footnotesize
\begin{adjustbox}{width=13.5cm}
\begin{tabular}{|l|l|c|c|c|c|c|c|c|}
\hline
\multicolumn{2}{|l|}{\multirow{2}{*}{\shortstack{4000 actions,\\ avg. over 100 instances}}} & \multirow{2}{*}{h} & \multirow{2}{*}{\# traj.} & \multirow{2}{*}{\shortstack{avg. cycle\\ time (s)}} & \multirow{2}{*}{\shortstack{st. dev.\\ (s)}} & \multirow{2}{*}{\shortstack{gain from\\ Rand. (\%)}} & \multirow{2}{*}{\shortstack{\# op.\\ tasks}} & \multirow{2}{*}{\shortstack{time \\(s)/action}} \\
\multicolumn{2}{|l|}{} &  &  &  &  &  &  &  \\ \hline
\multirow{32}{*}{\thead[cl]{No\\ learning}} & Random & - & - & 39.03 & 0.17 & - & 144.61 & $1.16\mathrm{e}{-4}$ \\ \cline{2-9} 
 & COL & - & - & 38.25 & 0.19 & 1.99 & 143.11 & $1.14\mathrm{e}{-4}$ \\ \cline{2-9} 
 & Class-based & - & - & 37.07 & 0.19 & 5.02 & 140.64 & $1.16\mathrm{e}{-4}$ \\ \cline{2-9} 
 & SL & - & - & 36.14 & 0.20 & 7.42 & 143.90 & $1.16\mathrm{e}{-4}$ \\ \cline{2-9} 
 & \multirow{7}{*}{\thead[cl]{SL + \\rollouts}} & 5 & - & 34.34 & 0.16 & 12.02 & 151.76 & $2.44\mathrm{e}{-3}$ \\ \cline{3-9} 
 &  & 10 & - & 33.86 & 0.17 & 13.26 & 182.32 & $3.93\mathrm{e}{-3}$ \\ \cline{3-9} 
 &  & 20 & - & 33.39 & 0.15 & 14.46 & 181.87 & $6.82\mathrm{e}{-3}$ \\ \cline{3-9} 
 &  & 30 & - & 33.09 & 0.16 & 15.21 & 181.19 & $9.80\mathrm{e}{-3}$ \\ \cline{3-9} 
 &  & 40 & - & 32.95 & 0.16 & 15.59 & 180.34 & $1.27\mathrm{e}{-2}$ \\ \cline{3-9} 
 &  & 50 & - & 32.89 & 0.16 & 15.75 & 181.39 & $1.57\mathrm{e}{-2}$ \\ \cline{3-9} 
 &  & 60 & - & 32.85 & 0.17 & 15.83 & 180.85 & $1.86\mathrm{e}{-2}$ \\ \cline{2-9} 
 & \multirow{21}{*}{MCTS} & 5 & 500 & 33.99 & 0.16 & 12.92 & 146.72 & $4.04\mathrm{e}{-1}$ \\ \cline{3-9} 
 &  & 10 & 500 & 32.97 & 0.18 & 15.54 & 190.10 & 1.00 \\ \cline{3-9} 
 &  & 20 & 500 & 32.02 & 0.14 & 17.95 & 201.33 & 2.06 \\ \cline{3-9}
 &  & 30 & 500 & 31.46 & 0.16 & 19.41 & 202.84 & 3.43 \\ \cline{3-9}
 &  & 30 & 5 & 32.75 & 0.16 & 16.10 & 185.26 & $4.61\mathrm{e}{-2}$ \\ \cline{3-9} 
 &  & 30 & 10 & 32.48 & 0.16 & 16.78 & 189.25 & $9.00\mathrm{e}{-2}$ \\ \cline{3-9} 
 &  & 30 & 20 & 32.26 & 0.14 & 17.34 & 192.14 & $1.75\mathrm{e}{-1}$ \\ \cline{3-9} 
 &  & 30 & 50 & 31.96 & 0.16 & 18.13 & 195.66 & $4.17\mathrm{e}{-1}$ \\ \cline{3-9} 
 &  & 30 & 75 & 31.85 & 0.15 & 18.39 & 198.45 & $6.12\mathrm{e}{-1}$ \\ \cline{3-9} 
 &  & 30 & 100 & 31.74 & 0.16 & 18.69 & 198.65 & $8.04\mathrm{e}{-1}$ \\ \cline{3-9} 
 &  & 30 & 200 & 31.60 & 0.15 & 19.05 & 201.58 & 1.54 \\ \cline{3-9} 
 &  & 30 & 1000 & 31.39 & 0.13 & 19.58 & 203.66 & 6.04 \\ \cline{3-9} 
 &  & 30 & 2000 & 31.40 & 0.15 & 19.56 & 203.59 & 10.38 \\ \cline{3-9} 
 &  & 60 & 5 & 32.48 & 0.18 & 16.79 & 185.83 & $8.77\mathrm{e}{-2}$ \\ \cline{3-9}
 &  & 60 & 10 & 32.22 & 0.16 & 17.46 & 189.02 & $1.71\mathrm{e}{-1}$ \\ \cline{3-9}
 &  & 60 & 20 & 31.95 & 0.17 & 18.13 & 191.71 & $3.39\mathrm{e}{-1}$ \\ \cline{3-9}
 &  & 60 & 50 & 31.63 & 0.17 & 18.96 & 194.33 & $8.22\mathrm{e}{-1}$ \\ \cline{3-9}
 &  & 60 & 75 & 31.49 & 0.15 & 19.33 & 196.53 & 1.22 \\ \cline{3-9}
 &  & 60 & 100 & 31.40 & 0.16 & 19.56 & 197.98 & 1.61 \\ \cline{3-9}
 &  & 60 & 200 & 31.19 & 0.16 & 20.09 & 200.01 & 3.17 \\ \cline{3-9} 
 &  & 60 & 500 & 30.93 & 0.17 & 20.75 & 203.46 & 7.79 \\ \hline
 
\end{tabular}
\end{adjustbox}
\end{table}

\begin{table}[H]
\centering
\label{table:full_results_mcts5}
\caption{Results of the methods learned from MCTS(5)}
\footnotesize
\begin{adjustbox}{width=13.5cm}
\begin{tabular}{|l|l|c|c|c|c|c|c|c|}
\hline
\multicolumn{2}{|l|}{\multirow{2}{*}{\shortstack{4000 actions,\\ avg. over 100 instances}}} & \multirow{2}{*}{h} & \multirow{2}{*}{\# traj.} & \multirow{2}{*}{\shortstack{avg. cycle\\ time (s)}} & \multirow{2}{*}{\shortstack{st. dev.\\ (s)}} & \multirow{2}{*}{\shortstack{gain from\\ Rand. (\%)}} & \multirow{2}{*}{\shortstack{\# op.\\ tasks}} & \multirow{2}{*}{\shortstack{time \\(s)/action}} \\
\multicolumn{2}{|l|}{} &  &  &  &  &  &  &  \\ \hline
\multirow{21}{*}{MCTS(5)} & LP & - & - & 34.44 & 0.17 & 11.76 & 143.81 & $1.10\mathrm{e}{-3}$ \\ \cline{2-9} 
 & \multirow{7}{*}{\thead[cl]{LP +\\ rollouts}} & 5 & - & 33.95 & 0.18 & 13.01 & 153.13 & $1.47\mathrm{e}{-2}$ \\ \cline{3-9} 
 &  & 10 & - & 33.27 & 0.19 & 14.76 & 184.31 & $2.82\mathrm{e}{-2}$ \\ \cline{3-9} 
 &  & 20 & - & 32.53 & 0.16 & 16.66 & 184.49 & $5.56\mathrm{e}{-2}$ \\ \cline{3-9} 
 &  & 30 & - & 32.13 & 0.18 & 17.69 & 185.19 & $8.27\mathrm{e}{-2}$ \\ \cline{3-9} 
 &  & 40 & - & 31.92 & 0.17 & 18.22 & 184.72 & $1.10\mathrm{e}{-1}$ \\ \cline{3-9} 
 &  & 50 & - & 31.85 & 0.15 & 18.41 & 184.78 & $1.37\mathrm{e}{-1}$ \\ \cline{3-9} 
 &  & 60 & - & 31.78 & 0.14 & 18.58 & 184.83 & $1.62\mathrm{e}{-1}$ \\ \cline{2-9} 
 & \multirow{7}{*}{\thead[cl]{Supervised\\ MCTS}} & 30 & 5 & 32.64 & 0.17 & 16.38 & 181.26 & $5.10\mathrm{e}{-2}$ \\ \cline{3-9} 
 &  & 30 & 10 & 32.48 & 0.16 & 16.79 & 180.95 & $9.93\mathrm{e}{-2}$ \\ \cline{3-9} 
 &  & 30 & 20 & 32.34 & 0.13 & 17.15 & 181.26 & $1.92\mathrm{e}{-1}$ \\ \cline{3-9} 
 &  & 30 & 50 & 32.14 & 0.16 & 17.66 & 183.14 & $4.63\mathrm{e}{-1}$ \\ \cline{3-9} 
 &  & 30 & 75 & 32.03 & 0.17 & 17.93 & 184.03 & $6.73\mathrm{e}{-1}$ \\ \cline{3-9} 
 &  & 30 & 100 & 31.96 & 0.17 & 18.12 & 184.75 & $8.83\mathrm{e}{-1}$ \\ \cline{3-9} 
 &  & 30 & 200 & 31.81 & 0.16 & 18.50 & 186.28 &  1.71\\ \cline{2-9} 
 & \multirow{7}{*}{STS} & 30 & 5 & 32.24 & 0.15 & 17.41 & 174.27 & $1.06\mathrm{e}{-1}$ \\ \cline{3-9} 
 &  & 30 & 10 & 31.90 & 0.15 & 18.26 & 183.81 & $2.07\mathrm{e}{-1}$ \\ \cline{3-9} 
 &  & 30 & 20 & 31.82 & 0.16 & 18.49 & 186.22 & $4.01\mathrm{e}{-1}$ \\ \cline{3-9} 
 &  & 30 & 50 & 31.66 & 0.16 & 18.90 & 189.97 & $9.24\mathrm{e}{-1}$ \\ \cline{3-9} 
 &  & 30 & 75 & 31.56 & 0.16 & 19.16 & 192.88 & 1.34 \\ \cline{3-9} 
 &  & 30 & 100 & 31.51 & 0.17 & 19.27 & 193.83 & 1.74 \\ \cline{3-9} 
 &  & 30 & 200 & 31.43 & 0.16 & 19.49 & 196.76 & 3.32 \\ \hline
 
\end{tabular}
\end{adjustbox}
\end{table}

\begin{table}[H]
\centering
\label{table:full_results_mcts10}
\caption{Results of the methods learned from MCTS(10)}
\footnotesize
\begin{adjustbox}{width=13.5cm}
\begin{tabular}{|l|l|c|c|c|c|c|c|c|}
\hline
\multicolumn{2}{|l|}{\multirow{2}{*}{\shortstack{4000 actions,\\ avg. over 100 instances}}} & \multirow{2}{*}{h} & \multirow{2}{*}{\# traj.} & \multirow{2}{*}{\shortstack{avg. cycle\\ time (s)}} & \multirow{2}{*}{\shortstack{st. dev.\\ (s)}} & \multirow{2}{*}{\shortstack{gain from\\ Rand. (\%)}} & \multirow{2}{*}{\shortstack{\# op.\\ tasks}} & \multirow{2}{*}{\shortstack{time \\(s)/action}} \\
\multicolumn{2}{|l|}{} &  &  &  &  &  &  &  \\ \hline
\multirow{28}{*}{MCTS(10)} & LP & - & - & 34.28 & 0.16 & 12.18 & 143.73 & $1.10\mathrm{e}{-3}$ \\ \cline{2-9} 
 & \multirow{7}{*}{\thead[cl]{LP +\\ rollouts}} & 5 & - & 33.96 & 0.16 & 12.99 & 152.48 & $1.51\mathrm{e}{-2}$ \\ \cline{3-9} 
 &  & 10 & - & 33.19 & 0.15 & 14.97 & 184.2 & $2.90\mathrm{e}{-2}$ \\ \cline{3-9} 
 &  & 20 & - & 32.47 & 0.17 & 16.81 & 184.56 & $5.71\mathrm{e}{-2}$ \\ \cline{3-9} 
 &  & 30 & - & 32.04 & 0.17 & 17.91 & 185.84 & $8.51\mathrm{e}{-2}$ \\ \cline{3-9} 
 &  & 40 & - & 31.88 & 0.16 & 18.32 & 185.41 & $1.13\mathrm{e}{-1}$ \\ \cline{3-9} 
 &  & 50 & - & 31.75 & 0.16 & 18.66 & 185.75 & $1.40\mathrm{e}{-1}$ \\ \cline{3-9} 
 &  & 60 & - & 31.73 & 0.15 & 18.71 & 185.33 & $1.67\mathrm{e}{-1}$ \\ \cline{2-9} 
 & \multirow{7}{*}{\thead[cl]{Supervised\\ MCTS}} & 30 & 5 & 32.61 & 0.18 & 16.46 & 183.13 & $5.17\mathrm{e}{-2}$ \\ \cline{3-9} 
 &  & 30 & 10 & 32.40 & 0.16 & 16.99 & 183.2 & $1.01\mathrm{e}{-1}$ \\ \cline{3-9} 
 &  & 30 & 20 & 32.20 & 0.17 & 17.52 & 184.99 & $1.99\mathrm{e}{-1}$ \\ \cline{3-9} 
 &  & 30 & 50 & 31.99 & 0.16 & 18.05 & 186.91 & $4.83\mathrm{e}{-1}$ \\ \cline{3-9} 
 &  & 30 & 75 & 31.92 & 0.16 & 18.22 & 187.3 & $7.14\mathrm{e}{-1}$ \\ \cline{3-9} 
 &  & 30 & 100 & 31.85 & 0.19 & 18.39 & 189.48 & $9.49\mathrm{e}{-1}$ \\ \cline{3-9} 
 &  & 30 & 200 & 31.74 & 0.16 & 18.68 & 190.75 & 1.87 \\ \cline{2-9} 
 & \multirow{14}{*}{STS} & 30 & 5 & 32.03 & 0.14 & 17.93 & 178.14 & $1.11\mathrm{e}{-1}$ \\ \cline{3-9} 
 &  & 30 & 10 & 31.79 & 0.15 & 18.57 & 184.86 & $2.17\mathrm{e}{-1}$ \\ \cline{3-9} 
 &  & 30 & 20 & 31.66 & 0.16 & 18.89 & 189.24 & $4.19\mathrm{e}{-1}$ \\ \cline{3-9} 
 &  & 30 & 50 & 31.49 & 0.16 & 19.31 & 192.66 & $9.87\mathrm{e}{-1}$ \\ \cline{3-9} 
 &  & 30 & 75 & 31.44 & 0.15 & 19.44 & 193.61 & 1.44 \\ \cline{3-9} 
 &  & 30 & 100 & 31.42 & 0.15 & 19.51 & 195.35 & 1.88 \\ \cline{3-9} 
 &  & 30 & 200 & 31.33 & 0.15 & 19.72 & 197.44 & 3.61 \\ \cline{3-9}
 &  & 60 & 5 & 31.66 & 0.16 & 18.89 & 179.28 & $2.28\mathrm{e}{-1}$ \\ \cline{3-9} 
 &  & 60 & 10 & 31.36 & 0.16 & 19.66 & 186.19 & $4.52\mathrm{e}{-1}$ \\ \cline{3-9} 
 &  & 60 & 20 & 31.24 & 0.17 & 19.97 & 188.28 & $8.90\mathrm{e}{-1}$ \\ \cline{3-9} 
 &  & 60 & 50 & 31.03 & 0.14 & 20.51 & 192.50 & 2.17 \\ \cline{3-9} 
 &  & 60 & 75 & 30.93 & 0.15 & 20.76 & 193.55 & 3.26 \\ \cline{3-9} 
 &  & 60 & 100 & 30.88 & 0.16 & 20.88 & 194.23 & 4.29 \\ \cline{3-9}
 &  & 60 & 200 & 30.78 & 0.14 & 21.15 & 197.11 & 8.64 \\ \hline
 
\end{tabular}
\end{adjustbox}
\end{table}

\begin{table}[H]
\centering
\label{table:full_results_mcts20}
\caption{Results of the methods learned from MCTS(20)}
\footnotesize
\begin{adjustbox}{width=13.5cm}
\begin{tabular}{|l|l|c|c|c|c|c|c|c|}
\hline
\multicolumn{2}{|l|}{\multirow{2}{*}{\shortstack{4000 actions,\\ avg. over 100 instances}}} & \multirow{2}{*}{h} & \multirow{2}{*}{\# traj.} & \multirow{2}{*}{\shortstack{avg. cycle\\ time (s)}} & \multirow{2}{*}{\shortstack{st. dev.\\ (s)}} & \multirow{2}{*}{\shortstack{gain from\\ Rand. (\%)}} & \multirow{2}{*}{\shortstack{\# op.\\ tasks}} & \multirow{2}{*}{\shortstack{time \\(s)/action}} \\
\multicolumn{2}{|l|}{} &  &  &  &  &  &  &  \\ \hline
\multirow{21}{*}{MCTS(20)} & LP & - & - & 34.32 & 0.18 & 12.06 & 144.24 & $1.10\mathrm{e}{-3}$ \\ \cline{2-9} 
 & \multirow{7}{*}{\thead[cl]{LP +\\ rollouts}} & 5 & - & 33.96 & 0.19 & 12.98 & 152.63 & $1.55\mathrm{e}{-2}$ \\ \cline{3-9} 
 &  & 10 & - & 33.21 & 0.19 & 14.92 & 184.11 & $2.99\mathrm{e}{-2}$ \\ \cline{3-9} 
 &  & 20 & - & 32.47 & 0.16 & 16.80 & 185.60 & $5.88\mathrm{e}{-2}$ \\ \cline{3-9} 
 &  & 30 & - & 32.06 & 0.16 & 17.86 & 185.49 & $8.77\mathrm{e}{-2}$ \\ \cline{3-9} 
 &  & 40 & - & 31.90 & 0.16 & 18.28 & 185.26 & $1.16\mathrm{e}{-1}$ \\ \cline{3-9} 
 &  & 50 & - & 31.79 & 0.16 & 18.56 & 185.20 & $1.44\mathrm{e}{-1}$ \\ \cline{3-9} 
 &  & 60 & - & 31.74 & 0.17 & 18.67 & 185.25 & $1.72\mathrm{e}{-1}$ \\ \cline{2-9} 
 & \multirow{7}{*}{\thead[cl]{Supervised\\ MCTS}} & 30 & 5 & 32.59 & 0.17 & 16.51 & 183.99 & $5.28\mathrm{e}{-2}$ \\ \cline{3-9} 
 &  & 30 & 10 & 32.40 & 0.15 & 16.98 & 185.70 & $1.03\mathrm{e}{-1}$ \\ \cline{3-9} 
 &  & 30 & 20 & 32.20 & 0.16 & 17.49 & 185.96 & $2.01\mathrm{e}{-1}$ \\ \cline{3-9} 
 &  & 30 & 50 & 31.99 & 0.18 & 18.05 & 188.58 & $4.87\mathrm{e}{-1}$ \\ \cline{3-9} 
 &  & 30 & 75 & 31.89 & 0.15 & 18.31 & 189.36 & $7.22\mathrm{e}{-1}$ \\ \cline{3-9} 
 &  & 30 & 100 & 31.84 & 0.15 & 18.43 & 190.65 & $9.55\mathrm{e}{-1}$ \\ \cline{3-9} 
 &  & 30 & 200 & 31.73 & 0.17 & 18.72 & 191.76 & 1.88 \\ \cline{2-9} 
 & \multirow{7}{*}{STS} & 30 & 5 & 32.01 & 0.16 & 18.00 & 179.06 & $1.16\mathrm{e}{-1}$ \\ \cline{3-9} 
 &  & 30 & 10 & 31.80 & 0.16 & 18.54 & 185.96 & $2.26\mathrm{e}{-1}$ \\ \cline{3-9} 
 &  & 30 & 20 & 31.64 & 0.15 & 18.94 & 188.62 & $4.39\mathrm{e}{-1}$ \\ \cline{3-9} 
 &  & 30 & 50 & 31.46 & 0.15 & 19.41 & 193.67 & 1.04 \\ \cline{3-9} 
 &  & 30 & 75 & 31.40 & 0.16 & 19.55 & 195.49 & 1.53 \\ \cline{3-9} 
 &  & 30 & 100 & 31.37 & 0.17 & 19.62 & 195.58 & 2.26 \\ \cline{3-9} 
 &  & 30 & 200 & 31.28 & 0.15 & 19.85 & 198.96 & 3.84 \\ \hline
 
\end{tabular}
\end{adjustbox}
\end{table}

\begin{table}[H]
\centering
\label{table:full_results_mcts30}
\caption{Results of the methods learned from MCTS(30)}
\footnotesize
\begin{adjustbox}{width=13.5cm}
\begin{tabular}{|l|l|c|c|c|c|c|c|c|}
\hline
\multicolumn{2}{|l|}{\multirow{2}{*}{\shortstack{4000 actions,\\ average on 100 instances}}} & \multirow{2}{*}{h} & \multirow{2}{*}{\# traj} & \multirow{2}{*}{\shortstack{avg cycle\\ time (s)}} & \multirow{2}{*}{\shortstack{st dev\\ (s)}} & \multirow{2}{*}{\shortstack{gain from\\ Rand. (\%)}} & \multirow{2}{*}{\shortstack{\# op.\\ tasks}} & \multirow{2}{*}{\shortstack{time \\(s)/action}} \\
\multicolumn{2}{|l|}{} &  &  &  &  &  &  &  \\ \hline
\multirow{21}{*}{MCTS(30)} & LP & - & - & 34.35 & 0.18 & 12.01 & 144.59 & $1.15\mathrm{e}{-3}$ \\ \cline{2-9} 
 & \multirow{7}{*}{\thead[cl]{LP +\\ rollouts}} & 5 & - & 33.96 & 0.17 & 12.99 & 152.44 & $1.61\mathrm{e}{-2}$ \\ \cline{3-9} 
 &  & 10 & - & 33.23 & 0.16 & 14.88 & 184.24 & $3.08\mathrm{e}{-2}$ \\ \cline{3-9} 
 &  & 20 & - & 32.49 & 0.18 & 16.77 & 184.15 & $6.07\mathrm{e}{-2}$ \\ \cline{3-9} 
 &  & 30 & - & 32.06 & 0.15 & 17.86 & 185.41 & $9.03\mathrm{e}{-2}$ \\ \cline{3-9} 
 &  & 40 & - & 31.88 & 0.18 & 18.31 & 184.94 & $1.20\mathrm{e}{-1}$ \\ \cline{3-9} 
 &  & 50 & - & 31.77 & 0.17 & 18.61 & 184.65 & $1.50\mathrm{e}{-1}$ \\ \cline{3-9} 
 &  & 60 & - & 31.74 & 0.16 & 18.69 & 183.84 & $1.79\mathrm{e}{-1}$ \\ \cline{2-9} 
 & \multirow{7}{*}{\thead[cl]{Supervised\\ MCTS}} & 30 & 5 & 32.63 & 0.16 & 16.41 & 184.76 & $5.25\mathrm{e}{-2}$ \\ \cline{3-9} 
 &  & 30 & 10 & 32.40 & 0.15 & 16.98 & 184.48 & $1.03\mathrm{e}{-1}$ \\ \cline{3-9} 
 &  & 30 & 20 & 32.22 & 0.16 & 17.44 & 185.36 & $2.01\mathrm{e}{-1}$ \\ \cline{3-9} 
 &  & 30 & 50 & 31.96 & 0.16 & 18.13 & 187.95 & $4.87\mathrm{e}{-1}$ \\ \cline{3-9} 
 &  & 30 & 75 & 31.91 & 0.18 & 18.25 & 189.93 & $7.23\mathrm{e}{-1}$ \\ \cline{3-9} 
 &  & 30 & 100 & 31.85 & 0.17 & 18.41 & 189.27 & $9.57\mathrm{e}{-1}$ \\ \cline{3-9} 
 &  & 30 & 200 & 31.73 & 0.16 & 18.71 & 191.26 & 1.89 \\ \cline{2-9} 
 & \multirow{7}{*}{STS} & 30 & 5 & 32.01 & 0.16 & 18.00 & 178.69 & $1.20\mathrm{e}{-1}$ \\ \cline{3-9} 
 &  & 30 & 10 & 31.78 & 0.16 & 18.59 & 184.68 & 18.59 \\ \cline{3-9} 
 &  & 30 & 20 & 31.62 & 0.18 & 19.00 & 188.97 & $4.56\mathrm{e}{-1}$ \\ \cline{3-9} 
 &  & 30 & 50 & 31.48 & 0.15 & 19.36 & 192.71 & 1.08 \\ \cline{3-9} 
 &  & 30 & 75 & 31.39 & 0.16 & 19.58 & 194.28 & 1.59 \\ \cline{3-9} 
 &  & 30 & 100 & 31.39 & 0.15 & 19.58 & 195.17 & 2.07 \\ \cline{3-9} 
 &  & 30 & 200 & 31.29 & 0.16 & 19.84 & 198.68 & 4.01 \\ \hline
 
\end{tabular}
\end{adjustbox}
\end{table}

\newpage
\bibliography{references.bib}

\end{document}